\documentclass[runningheads]{llncs}
\usepackage[T1]{fontenc}
\usepackage{graphicx}

\usepackage[table]{xcolor}
\usepackage{booktabs}
\usepackage{multirow}
\usepackage{amsmath}
\usepackage{graphicx}
\usepackage{bbding} %
\usepackage{pifont}
\usepackage{wasysym}
\usepackage{amssymb}

\usepackage{authblk}
\usepackage{hyperref}
\hypersetup{colorlinks=true,citecolor=blue}
\usepackage{graphicx,subfigure}

\newcommand\blfootnote[1]
{
	\begingroup
	\renewcommand\thefootnote{}
	\footnotetext{#1}%
	\endgroup
}

\def\ourdataset{\textit{TVMI3K}}
\def\ourmodel{\emph{TVNet}}

\newif\ifreview

\title{Trichomonas Vaginalis Segmentation in Microscope Images}

\begin{document}

\author{Lin Li\inst{1,2} \and
Jingyi Liu\inst{3} \and
Shuo Wang\inst{4} \and 
Xunkun Wang~\inst{1}~\Envelope \and
Tian-Zhu Xiang~\inst{5}~\Envelope
}

\authorrunning{Li et al.}

\institute{Qingdao Huajing Biotechnology CO., LTD \and
Ocean University of China, Qingdao, China \and
Qingdao University of Science and Technology, Qingdao, China \and 
ETH Z{\"u}rich, Z{\"u}rich, Switzerland \and
Inception Institute of Artificial Intelligence, Abu Dhabi, UAE \\
%\email{}
}

\maketitle  

\blfootnote{\Envelope~Co-corresponding authors (maksimljc@163.com,~tianzhu.xiang19@gmail.com). \\ First Author: Lin Li (ll198196@163.com).}

\begin{abstract}
Trichomoniasis is a common infectious disease with high incidence caused by the parasite Trichomonas vaginalis, increasing the risk of getting HIV in humans if left untreated.
Automated detection of Trichomonas vaginalis from microscopic images can provide vital information for diagnosis of trichomoniasis. 
However, accurate Trichomonas vaginalis segmentation (TVS) is a challenging task due to the high appearance similarity between the Trichomonas and other cells (\textit{e.g.}, leukocyte), the large appearance variation caused by their motility, and, most importantly, the lack of large-scale annotated data for deep model training.
To address these challenges, we elaborately collected the first large-scale Microscopic Image dataset of Trichomonas Vaginalis, named \textbf{\ourdataset}, which consists of 3,158 images covering Trichomonas of various appearances in diverse backgrounds, with high-quality annotations including object-level mask labels, object boundaries, and challenging attributes. 
Besides, we propose a simple yet effective baseline, termed \textbf{\ourmodel}, to automatically segment Trichomonas from microscopic images, including high-resolution fusion and foreground-background attention modules. 
Extensive experiments demonstrate that our model achieves superior segmentation performance and outperforms various cutting-edge object detection models both quantitatively and qualitatively, making it a promising framework to promote future research in TVS tasks.

\end{abstract}

\section{Introduction}
Trichomoniasis (or ``trich''), caused by infection with a motile, flagellated protozoan parasite called Trichomonas vaginalis (TV), is likely the most common, non-viral sexually transmitted infection (STI) worldwide.
According to statistics, there are more than 160 million new cases of trichomoniasis in the world each year, with a similar probability of males and females~\cite{harp2011trichomoniasis,vos2016global}.
A number of studies have shown that Trichomonas vaginalis infection is associated with an increased risk of infection with several other STIs, including human papillomavirus (HPV) and human immunodeficiency virus (HIV)~\cite{workowski2012sexually}. The high prevalence of Trichomonas vaginalis infection globally and the frequency of co-infection with other STIs make trichomoniasis a compelling public health concern.

Automatic Trichomonas vaginalis segmentation (TVS) is crucial to the diagnosis of Trichomoniasis.
Recently, deep learning methods have been widely used for medical image segmentation~\cite{hesamian2019deep} and made significant progress, such as brain region and tumor segmentation~\cite{havaei2017brain,zhao2018deep}, liver and tumor segmentation~\cite{zhou2018unet++,tang2020two}, polyp segmentation~\cite{brandao2017fully,fan2020pra,ji2021progressively}, lung infection segmentation~\cite{fan2020inf,liu2021covid} and cell segmentation~\cite{ronneberger2015u,li2021mvdi25k,zhang2021multi}. 
Most of these methods are based on the encoder-decoder framework, such as U-Net\cite{ronneberger2015u} and its variants~\cite{siddique2021u} (\textit{e.g.}, U-Net++~\cite{zhou2018unet++}, Unet 3+~\cite{huang2020unet}), and PraNet~\cite{fan2020pra}, or are inspired by some commonly-used natural image segmentation models, \textit{e.g.}, fully convolutional networks (FCN)~\cite{zhao2018deep} and DeepLab~\cite{tang2020two}. 
These works have shown great potentials in the segmentation of various organs and lesions from different medical imaging modalities. 
To our knowledge, however, deep learning techniques have not yet been well-studied and applied for TVS in microscope images, due to three key factors: 
1) The large variation in morphology (\textit{e.g.}, size, appearance and shape) of the Trichomonas is challenging for detection. Besides, Trichomonas are often captured out of focus (blurred appearance) or under occlusion due to their motility, which aggregates the difficulty of accurate segmentation.
2) The high appearance similarity between Trichomonas and other cells (\textit{e.g.}, leukocyte) makes them easily confused with complex surroundings. 
Most importantly, 3) the lack of large-scale annotated data restricts the performance of deep models that rely on sufficient training data, thereby hindering further research in this field.
It is worth noting that the above factors also reflect the clear differences in object segmentation between the microscope images of Trichomonas in our work and conventional cells (\textit{e.g.}, HeLa cells~\cite{ronneberger2015u} and blood cells~\cite{li2021robust}). 
Furthermore, we noted that recently Wang et al.~\cite{wang2021trichomonas} proposed a two-stage model for video-based Trichomonas vaginalis detection, which utilizes video motion cues (\textit{e.g.}, optical flow) to greatly reduce the detection difficulty. The difference is that this work focuses on image-based Trichomonas vaginalis detection, without motion information, which increases the difficulty of detection. As we know, no one has set foot on this field so far. 
Hence, accurate TV segmentation remains a challenging and under-explored task.

To address above issues, we first elaborately construct a novel large-scale microscope images dataset exclusively designed for Trichomonas Vaginalis segmentation, named \textit{TVMI3K}. 
Moreover, we develop a simple but effective deep neural network, termed \textit{TVNet}, for TVS. 
In a nutshell, our main contributions are threefold: (1) We carefully collect \textbf{\textit{TVMI3K}, a large-scale dataset for TVS}, which consists of 3,158 microscopic images covering Trichomonas of various appearances in diverse backgrounds, with high-quality annotations of object-level labels, object boundaries and challenging attributes. To our knowledge, this is the first large-scale dataset for TVS that can serve as a catalyst for promoting further research in this field in the deep learning era. (2) We proposed \textbf{a novel deep neural network, termed \textit{TVNet}}, which enhances high-level feature representations with edge cues in a high-resolution fusion (HRF) module and then excavates object-critical semantics based on foreground-background attention (FBA) module under the guidance of coarse location map for accurate prediction. (3) Extensive experiments show that our method achieves superior performance and outperforms various cutting-edge segmentation models both quantitatively and qualitatively, making it a promising solution to the TVS task. 
The dataset, results and models will be publicly available at: \href{https://github.com/CellRecog/cellRecog}{https://github.com/CellRecog/cellRecog}.

\begin{figure}[t]
    \centering
    \includegraphics[width=.95\textwidth]{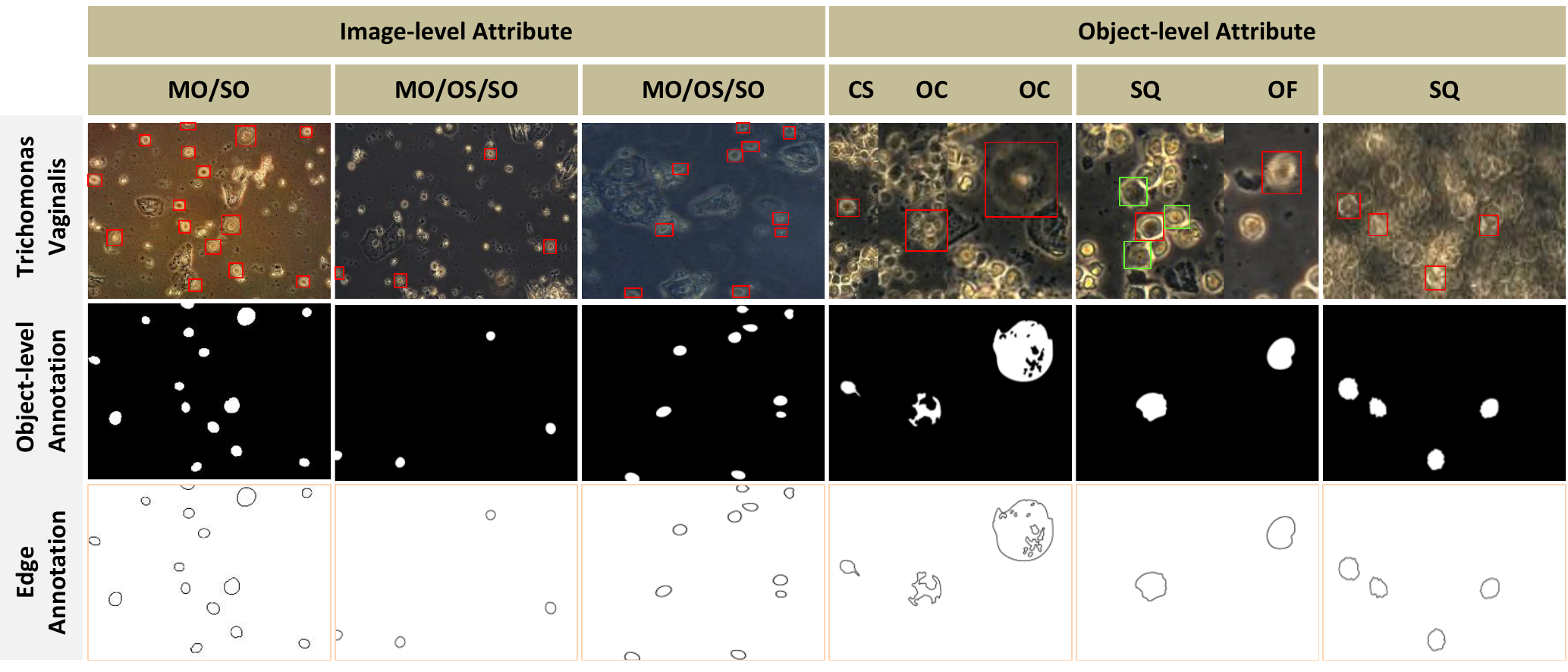}
    \caption{Various examples of our proposed \ourdataset. We provide different annotations, including object-level masks, object edges and challenging attributes. We use red boxes to mark Trichomonas and green boxes to mark leukocyte on images for better visualization. Leukocyte shows high similarity with Trichomonas.}
    \label{sample02}
\end{figure}

\section{Proposed Dataset}

To facilitate the research of TVS in deep learning era, we develop the \textit{TVMI3K} dataset, which is carefully collected to cover TV of various appearances in diverse challenging surroundings, \textit{e.g.}, large morphological variation, occlusion and background distractions. Examples can be seen in Fig.~\ref{sample02}.

\subsection{Data Collection}

To construct the dataset, we first collect 80 videos of Trichomonas samples with resolution 3088$\times$2064 from more than 20 cases over seven months. The phase-contrast microscopy is adopted for video collection, which captures samples clearly, even unstained cells, making it more suitable for sample imaging and microscopy evaluation than ordinary light microscopy. We then extract images from these videos to build our \textit{TVMI3K} dataset, which finally contains 3,158 microscopic images (2,524 Trichomonas and 634 background images). 
We find that videos are more beneficial for data annotation, because objects can be identified and annotated accurately according to the motion of Trichomonas. Thus, even unfocused objects can also be labeled accurately. 
Besides, to avoid data selection bias, we collect 634 background images to enhance generalization ability of models.
It should be noted that images are collected by the microscope devices independently, and we do not collect any patient information, thus the dataset is free from copyright and loyalties.

High-quality annotations are critical for deep model training~\cite{li2021systematic}. During the labeling process, 5 professional annotators are divided into two groups for annotation and cross-validation is conducted between each group to guarantee the quality of annotation. We label each image with accurately object-level masks, object edges and challenging attributes, \textit{e.g.}, occlusions and complex shapes. 
Attribute descriptions are shown in Tab.~\ref{Tab:Att}.

\begin{table}[t]
\caption{Attribute Descriptions of Our \ourdataset~Dataset}
\centering
\label{Tab:Att}
\begin{tabular}{cp{1.3cm}<{\centering}l}
%\rowcolor{gray!15}
\toprule
 & Attr.  &Descriptions  \\ 
\midrule

\multirow{3}*{
\rotatebox{90}{Image-}}
\multirow{3}*{
\rotatebox{90}{level}}
& \textbf{MO} & \textit{Multiple Objects}. Number of objects in each image $\geq$ 2 \\ % dense objects
& \textbf{SO} & \textit{Small Objects}. The ratio of object area to image area $\leq$ 0.1 \\
& \textbf{OV} & \textit{Out of view}. Incomplete objects clipped by image boundary \\

\hline 

\multirow{4}*{
\rotatebox{90}{Object-}}
\multirow{4}*{
\rotatebox{90}{level}}
& \textbf{CS} & \textit{Complex Shape}. In diverse shapes with tiny parts (\textit{e.g.}, flagella)  \\
& \textbf{OC} & \textit{Occlusions}. Object is partially obscured by surroundings \\
& \textbf{OF} & \textit{Out-of-focus}. Ghosting due to poor focus \\
& \textbf{SQ} & \textit{Squeeze}. The object appearance changes when squeezed \\
 
\bottomrule
\end{tabular}
\end{table}

\subsection{Dataset Features}

~~~~~$\bullet$ \emph{Image-level Attributes.} As listed in Tab.~\ref{Tab:Att}, our data is collected with several image-level attributes, \textit{i.e.}, multiple objects (MO), small object (SO) and out-of-view (OV), which are mostly caused by Trichomonas size, shooting distance and range. 
According to statistics, each image contains $3$ objects averagely, up to $17$ objects. The size distribution ranges from $0.029\%$ to $1.179\%$, with an average of $0.188\%$, indicating that it belongs to a dataset for tiny object detection.

$\bullet$ \emph{Object-level Attributes.} Trichomonas objects show various attributes, including complex shapes (CS), occlusions (OC), out-of-focus (OF) and squeeze (SQ), which are mainly caused by the growth and motion of trichomonas, highly background distraction (\textit{e.g}. other cells) and the shaking of acquisition devices. 
These attributes lead to large morphological differences in Trichomonas, which increase the difficulty of detection. In addition, there is a high similarity in appearance between Trichomonas and leukocyte, which can easily confuse the detectors to make false detections. Examples and the details of object-level attributes are shown in Fig.~\ref{sample02} and Tab.~\ref{Tab:Att} respectively.

$\bullet$ \emph{Dataset Splits.} To provide a large amount of training data for deep models, we select 60\% of videos as training set and 40\% as test set. In order to be consistent with practical applications, we select data according to the chronological order of the sample videos from each case instead of random selection, that is, the data collected first for each case is used as the training set, and the data collected later is used as the test set.
Thus, the dataset is finally split into 2,305 images for training and 853 images for testing respectively. Note that 290 background images in the test set are not included in our test experiments.

\section{Method}
\label{method}

\begin{figure}[t]
    \centering    
    \includegraphics[width=0.95\textwidth]{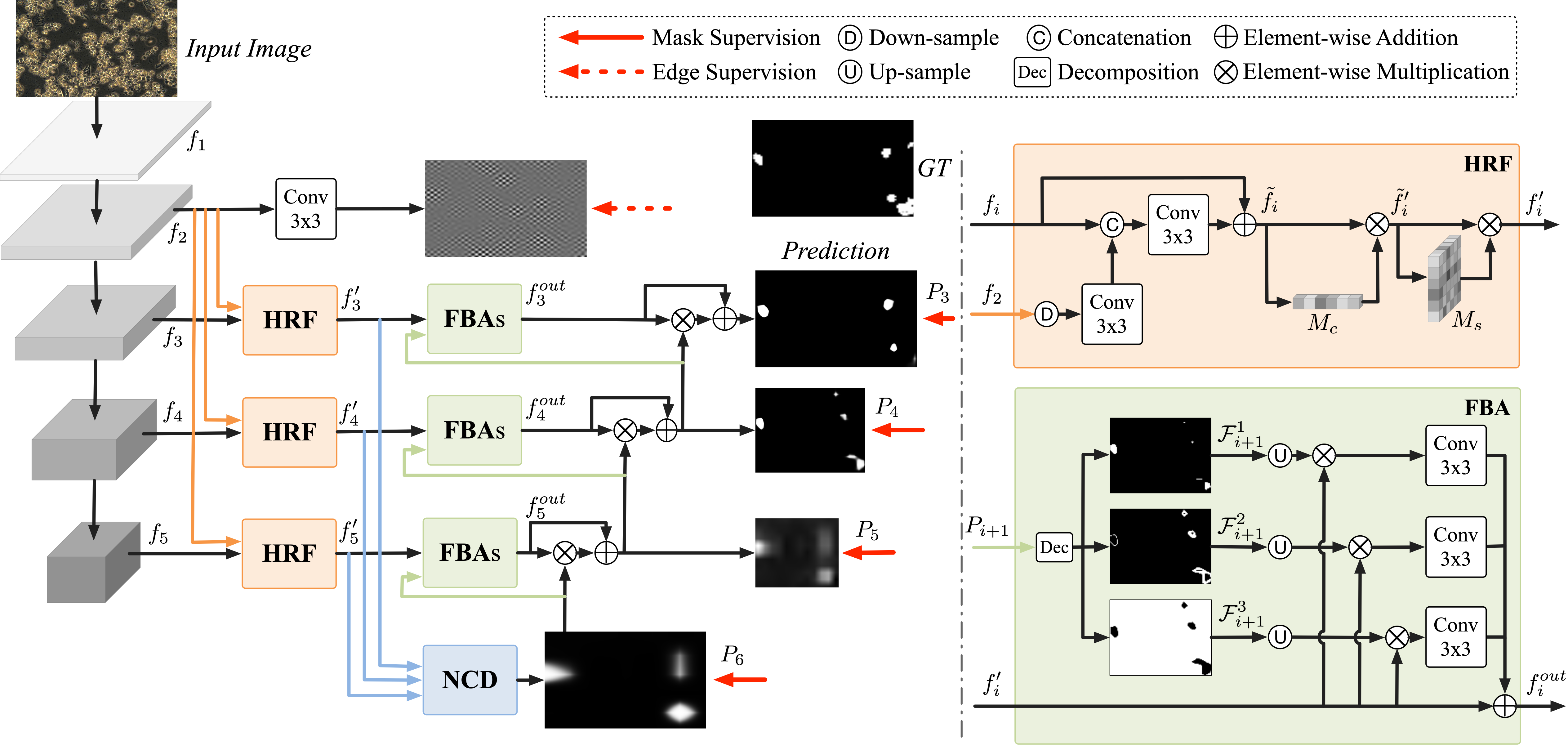}
     \caption{Overview of our proposed \ourmodel, which consists of high-resolution fusion (HRF) module and foreground-background attention (FBA) module. See $\S$~\ref{method} for details.}
    \label{net12}
\end{figure}

\subsection{Overview}
Fig.~\ref{net12} illustrates the overall architecture of the proposed \ourmodel. 
Specifically, for an input Trichomonas image $I$, the backbone network Res2Net~\cite{gao2019res2net} is adopted to extract five levels of features $\{f_i\}_{i=1}^5$. Then we further explore the high-level features (\textit{i.e.}, $f_3\sim f_5$ in our model) to effectively learn object feature representation and produce object prediction. As shown in Fig.~\ref{net12}, we first design a high-resolution fusion (HRF) module to enhance feature representation by integrating high-resolution edge features and get three refined features ($\{f_i^{'}\}_{i=3}^5$). 
Next, we introduce the neighbor connection decoder (NCD)~\cite{fan2021concealed}, a new cutting-edge decoder component which improves the partial decoder component~\cite{wu2019cascaded} with neighbor connection, to aggregate these three refined features and generate the initial prediction map $P_6$. It can capture the relatively coarse location of objects so as to guide the following object prediction. Specially, $\{P_i\}_{i=3}^5$ represents the prediction map of the $i$ layer, which corresponds to the $i$-layer in $\{f_i^{'}\}_{i=3}^5$.
Finally, we propose a foreground-background attention (FBA) module to excavate object critical cues by exploring semantic differences between foreground and background and then accurately predict object masks in a progressive manner.

\subsection{High-resolution Fusion Module}

With the deepening of the neural network, local features, such as textures and edges, are gradually diluted, which may reduce the model's capability to learn the object structure and boundaries.
Considering too small receptive field and too much redundant information of $f_1$ feature, in this paper, we design a high-resolution fusion (HRF) module which adopts the the low-level $f_2$ feature to supplement local details for high-level semantic features and boost feature extraction for segmentation. To force the model to focus more on the object edge information, we first feed feature $f_2$ into a $3\times 3$ convolutional layer to explicitly model object boundaries with the ground-truth edge supervision. 
Then we integrate the low-level feature $f_2$ with the high-level features ($f_3\sim f_5$) to enhance representation with channel and spatial attention operations~\cite{woo2018cbam,chen2017sca}, denoted as:
\begin{equation}
\left\{
\begin{aligned}
\Tilde{f}_i & = f_i + \mathcal{G} \left( Cat(f_i;~ \mathcal{G}(\delta_\downarrow^2 (f_2))) \right), i\in \{3,...,5\}, \\
\Tilde{f}_i^{'} & = M_c(\Tilde{f}_i) \otimes \Tilde{f}_i, \\
f_i^{'} & = M_s(\Tilde{f}_i^{'}) \otimes \Tilde{f}_i^{'}, 
\end{aligned}
\right.
\end{equation}
where $\delta_\downarrow^2(\cdot)$ denotes $\times 2$ down-sampling operation. $\mathcal{G}(\cdot)$ is a $3\times 3$ convolution layer, and $Cat$ is concatenation operation. $M_c(\cdot)$ and $M_s(\cdot)$ represent channel attention and spatial attention respectively. $\otimes$ is the element-wise multiplication operation. After that, a 1$\times$1 convolution is used to adjust the number of channels to ensure the consistent output of each HRF.

\subsection{Foreground-Background Attention Module} 
To produce accurate prediction from the fused feature ($f_3^{'}\sim f_5^{'}$) progressively in the decoder, we devise a foreground-background attention (FBA) module to filtrate and enhance object related features using region sensitive map derived from the coarse prediction of the previous layer. 
For the fused feature $\{f_i^{'}\}_{i=3}^5$, we first decompose the previous initial prediction $P_{i+1}$ into three regions, \textit{i.e.}, strong foreground region ($\mathcal{F}_{i+1}^{1}$), weak foreground region ($\mathcal{F}_{i+1}^{2}$) and background region ($\mathcal{F}_{i+1}^3$). The decomposition process can refer to~\cite{sun2021deep}. Then each region is normalized into [0,1] as a region sensitive map to extract the corresponding features from $f_i^{'}$. 
Noted that $\mathcal{F}^{1}$ provides the location information, and $\mathcal{F}^{2}$ contains the object edge/boundary information. The $\mathcal{F}^3$ denotes the remaining region. 

After that, the region sensitive features can be extracted from the fused feature by an element-wise multiplication operation with up-sampled region sensitive maps followed by a 3$\times$3 convolution layer. Next, these features are aggregated by an element-wise summation operation with a residual connection. It can be denoted as: 
\begin{equation}
f_{i}^{out} = \sum\nolimits_{k=1}^{3}
\mathcal{G} (\delta_\uparrow^2 (\mathcal{F}_{i+1}^{k}) \otimes f_i^{'}) + f_i^{'}
\end{equation}
where $\delta_\uparrow^2(\cdot)$ denotes $\times 2$ up-sampling operation. 
Inspired by~\cite{fan2021concealed}, FBA can also be cascaded multiple times to gradually refine the prediction. For more details please refer to the \textit{Supp.} In this way, FBA can differentially handle regions with different properties and explore region-sensitive features, thereby strengthening foreground features and reducing background distractions.

\section{Experiments and Results}

\subsection{Experimental Settings}

\textbf{Baselines and Metrics.} We compare our \ourmodel~with 9 state-of-the-art medical/natural image segmentation methods, including U-Net++~\cite{zhou2019unetplusplus}, SCRN~\cite{9010954}, U$^2$Net~\cite{qin2020u2}, F$^3$Net~\cite{f3net}, PraNet~\cite{fan2020pra}, SINet~\cite{SINet}, MSNet~\cite{MSNet}, SANet~\cite{wei2021shallow} and SINet-v2~\cite{fan2021concealed}. We collect the source codes of these models and re-train them on our proposed dataset.
We adopt 7 metrics for quantitative evaluation using the toolboxes provided by~\cite{fan2020pra} and~\cite{wei2021shallow}, including structural similarity measure ($S_\alpha$, $\alpha$ = $0.5$)~\cite{fan2017structure}, enhanced alignment measure ($E_\phi^{max}$)~\cite{fan2018enhanced}, $F_\beta$ measure ($F_\beta^w$ and $F_\beta^{mean}$)~\cite{margolin2014evaluate}, mean absolute error (MAE, $\mathcal{M}$)~\cite{perazzi2012saliency}, Sorensen-Dice coefficient (mean Dice, mDice) and intersection-over-union (mean IoU, mIoU)~\cite{fan2020pra}.

\noindent\textbf{Training Protocols.}
We adopt the standard binary cross entropy (BCE) loss for edge supervision, and the weighted IoU loss~\cite{f3net} and the weighted BCE loss~\cite{f3net} for object mask supervision. 
During the training stage, the batch size is set to 20. The network parameters are optimized by Adam optimizer~\cite{kingma2014adam} with an initial learning rate of 0.05, a momentum of 0.9 and a weight decay of 5e-4. 
Each image is resized to 352$\times$352 for network input. The whole training time is about 2 hours for 50 epochs on a NVIDIA GeForce RTX 2080Ti GPU.

\begin{table*}[t]%[width=.75\linewidth]
\caption{Quantitative comparison on our \ourdataset~dataset. ``$\uparrow$'' indicates the higher the score the better. ``$\downarrow$'' denotes the lower the score the better.}
\centering
\label{tbl1}
% \renewcommand{\arraystretch}{1.0}
% \setlength\tabcolsep{3.8pt}
% \scriptsize
\begin{tabular}{l|c||c|c|c|c|c|c|c}
  \toprule
  \rowcolor{gray!25}
   Methods & Pub.
               &$S_\alpha\uparrow$ 
              &$E_\phi^{max}\uparrow$    
              &~~$F_\beta^w\uparrow$~~ 
              &$F_\beta^{mean}\uparrow$
              &$\mathcal{M}\downarrow$ 
              &mDice$\uparrow$ 
              &mIoU$\uparrow$  
               \\
  \midrule
     UNet++~\cite{zhou2019unetplusplus} & TMI19
         &0.524   
         &0.731 
         &0.069   
         &0.053    
         &0.006
         &0.004     
         &0.003  
        \\
         %\hline
     SCRN~\cite{9010954} &ICCV19
         &0.567   
         &0.789 
         &0.145   
         &0.254    
         &0.011
         &0.201    
         &0.135 
         \\
        % \hline
     U$^2$Net~\cite{qin2020u2} &PR20
         &0.607   
         &0.845 
         &0.209   
         &0.332    
         &0.013
         &0.301     
         &0.209  
         \\
         %\hline
     F$^3$Net~\cite{f3net}  &AAAI20
         &\textbf{0.637}   
         &0.809 
         &0.320   
         &0.377    
         &0.005
         &0.369    
         &0.265    
         \\
         %\hline
     PraNet~\cite{fan2020pra} &MICCAI20
         &0.623   
         &0.792 
         &0.300   
         &0.369    
         &0.006 
          &0.328    
          &0.230  
          \\
     %\hline
     SINet~\cite{SINet}  &CVPR20
         &0.492   
         &0.752 
         &0.010
         &0.113    
         &0.104
         &0.095    
        &0.061  
        \\  %\hline
     MSNet~\cite{MSNet} &MICCAI21
         &0.626   
         &0.786 
         &0.321   
         &0.378    
         &0.005
         &0.366    
         &0.268  
        \\  %\hline
     SANet~\cite{wei2021shallow}  &MICCAI21
         &0.612    
         &0.800 
         &0.289   
         &0.361    
         &0.006
         &0.338    
         &0.225 
        \\
  %\hline
  SINet-v2~\cite{fan2021concealed}  &PAMI21
         &0.621   
         &0.842 
         &0.309   
         &0.375    
         &0.005 
         &0.348    
         &0.245   
         \\
         \midrule
  \textbf{Ours} &-
         &0.635
         &\textbf{0.851}
         &\textbf{0.343 }  
         &\textbf{0.401}    
         &\textbf{0.004}
         &\textbf{0.376}  
         &\textbf{0.276}    
         \\
    \toprule
    \end{tabular}
\end{table*}

\subsection{Comparison with State-of-the-art}
\textbf{Quantitative comparison.}
Tab.~\ref{tbl1} shows the quantitative comparison between our proposed model and other competitors on our \textit{TVMI3K} dataset. It can be seen that our TVNet significantly outperforms all other competing methods on all metrics except $S_\alpha$ which is also on par with the best one. 
Particularly, our method achieves a performance gain of 2.2\% and 2.3\% in terms of $F_\beta^w$ and {\footnotesize $F_\beta^{mean}$}, respectively. 
This suggests that our model is a strong baseline for TVS.

\begin{figure}[t]
\centering
    \includegraphics[width=0.95\textwidth]{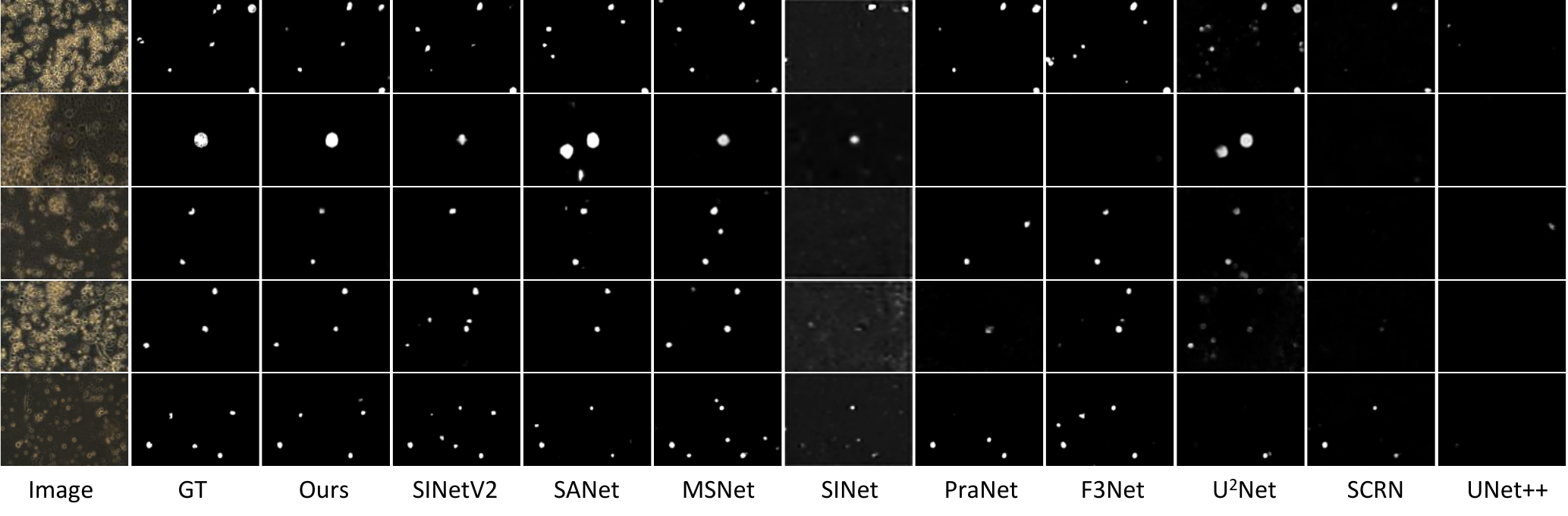}
     \caption{Visual comparison of different methods. Obviously, our method provides more accurate predictions than other competitors in various challenging scenarios.} 
    \label{netresult02}
\end{figure}

\noindent\textbf{Qualitative comparison.} Fig.~\ref{netresult02} shows some representative visual results of different methods. From those results, we can observe that TVNet can accurately locate and segment Trichomonas objects under various challenging scenarios, including cluttered distraction objects, occlusion, varied shape and similarity with other cells. 
In contrast, other methods often provide results with a considerable number of missed or false detection, or even failed detection.

\begin{table*}[t]
	\centering
	\caption{Ablation study for TVNet on the proposed \ourdataset~datasets.} 
	\label{Tab:ablation}
	\setlength\tabcolsep{4pt}
% 	\resizebox{1\linewidth}{!}{
		\begin{tabular}{cccc||cccccc}
			\toprule
			\rowcolor{gray!25}
			No. & Backbone & HRF &FBA 
			&$S_\alpha\uparrow$ 
              &~~$F_\beta^w\uparrow$~~ 
              &$F_\beta^{mean}\uparrow$
              &$\mathcal{M}\downarrow$ 	
              &mDice
              &mIoU\\
			\midrule
			a
			& \checkmark& &  
			&0.593
			&0.234
			&0.302
			&0.005
			&0.251
			&0.163
			\\ 
			\hline
			b& \checkmark &\checkmark&
			&0.619 
			&0.291
			&0.357 
			&0.005 
			&0.328
			&0.228
			\\
			\hline
			 
			c& \checkmark&&\checkmark 
			&0.623
			&0.258
			&0.330  
			&\textbf{0.004} 
			&0.271
			&0.185
			\\ 
			\hline

		 d& \checkmark&\checkmark&\checkmark
		 &\textbf{0.635}  
         &\textbf{0.343}
         &\textbf{0.401}    
         &\textbf{0.004}   
         &\textbf{0.376}
		&\textbf{0.276}\\ 
		
		\bottomrule
		\end{tabular}
% 	}
\end{table*}

\subsection{Ablation Study} 

\noindent\textbf{Effectiveness of HRF}. From Tab.~\ref{Tab:ablation}, we observe that the HRF module outperforms the baseline model with significant improvement, \textit{e.g.}, 2.6\%, 5.7\%, 5.7\%, 7.7\% and 6.5\% performance improvement in $S_\alpha$, $F_\beta^w$ $F_\beta^{mean}$, $mDice$ and $mIoU$ metrics, respectively. This shows the fusion of local features is beneficial for object boundary localization and segmentation. Note that the adopted explicit edge supervision facilitates the model to focus more on object boundaries and enhance the details of predictions.

\noindent\textbf{Effectiveness of FBA}. We further investigate the contribution of the FBA module. As can be seen in Tab.~\ref{Tab:ablation}, FBA improves the segmentation performance by 3\%, 2.4\% and 2.8\% in $S_\alpha$, $F_\beta^w$ and $F_\beta^{mean}$, respectively. FBA enables our model to excavate object-critical features and reduce background distractions, thus distinguishing TV objects accurately.

\noindent\textbf{Effectiveness of HRF \& FBA}. From Tab.~\ref{Tab:ablation}, the integration of HRF and FBA is generally better than other settings (a$\sim $c). Compared with the baseline, the performance gains are 1.2\%, 5.2\% and 4.4\% in $S_\alpha$, $F_\beta^w$ and $F_\beta^{mean}$ respectively. Besides, our \textit{TVNet} outperforms other recently proposed models, making it an effective framework that can help boost future research in TVS. 

\noindent\textbf{Model Complexity}. We observe that the number of parameters and FLOPs of the proposed model are $\sim$155M and $\sim$98GMac, respectively, indicating that there is room for further improvement, which is the focus of our future work.

\section{Conclusion}
This paper provides the first investigation for the segmentation of Trichomonas vaginalis in microscope images based on deep neural networks. To this end, we collect a novel large-scale, challenging microscope image dataset of TV called \ourdataset. 
Then, we propose a simple but effective baseline, \ourmodel, for accurately segmenting Trichomonas from microscope images. 
Extensive experiments demonstrate that our \ourmodel~outperforms other approaches.
We hope our study will offer the community an opportunity to explore more in this field.

%
% ---- Bibliography ----
%

\bibliographystyle{splncs04}
\bibliography{TVS-bib}

\newpage

\section{Appendix}
\vspace{-5pt}

\subsection{More Details of \textit{TVMI3K} Dataset}
\vspace{-20pt}

\begin{figure}
    \centering
    \includegraphics[width=.98\textwidth]{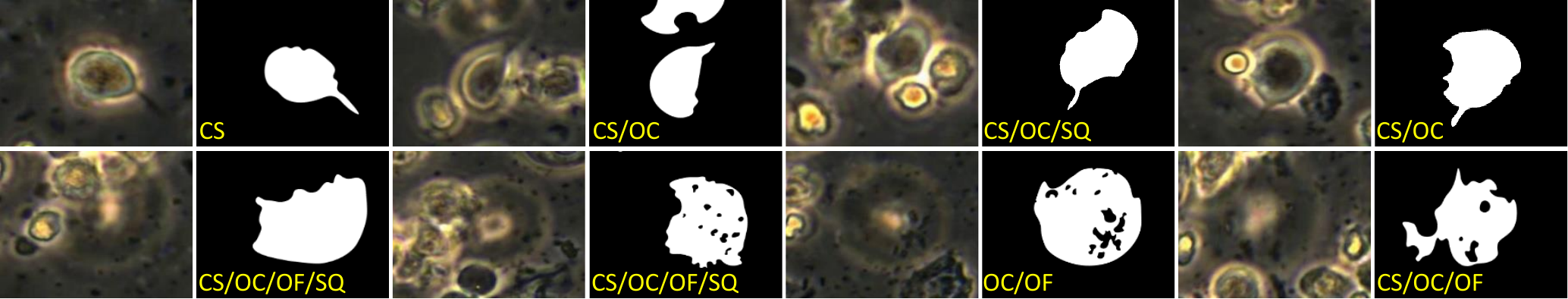}
    \caption{Examples of Object-level Attributes.}
    \label{FIG1}
\end{figure}
\vspace{-40pt}

\begin{figure*}[h]
\centering
\subfigure[Co-attribute Distribution]{
\begin{minipage}[c]{0.43\linewidth}
\centering
\includegraphics[width=0.58\textwidth]{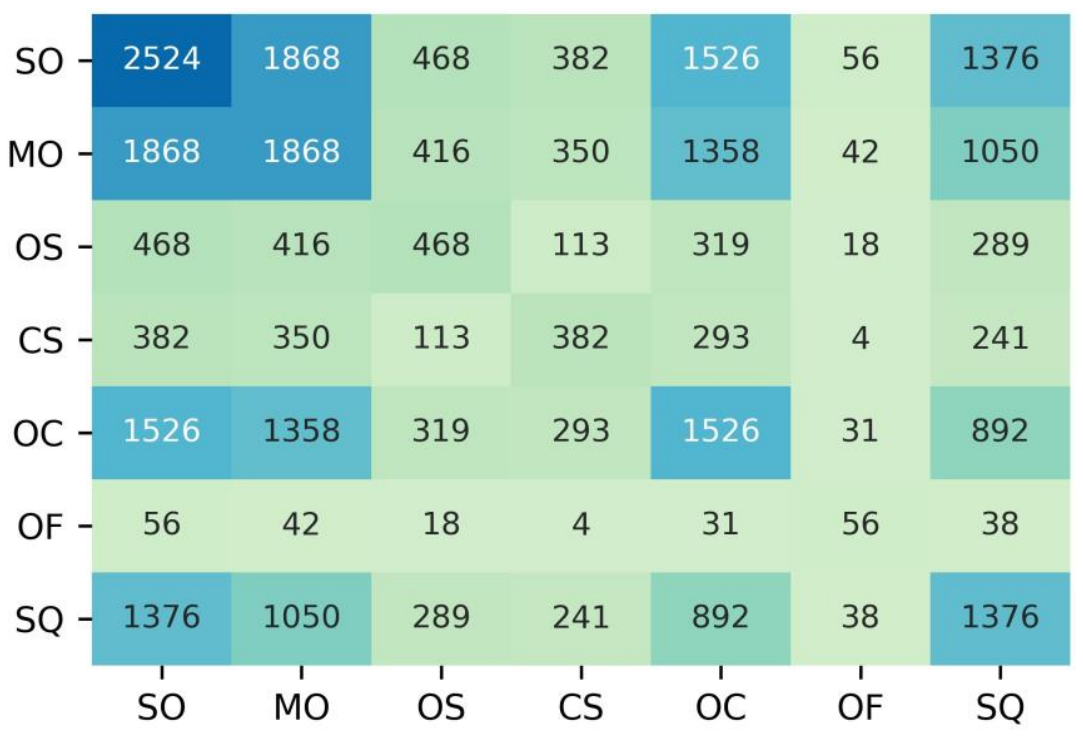} 
\includegraphics[width=0.4\textwidth]{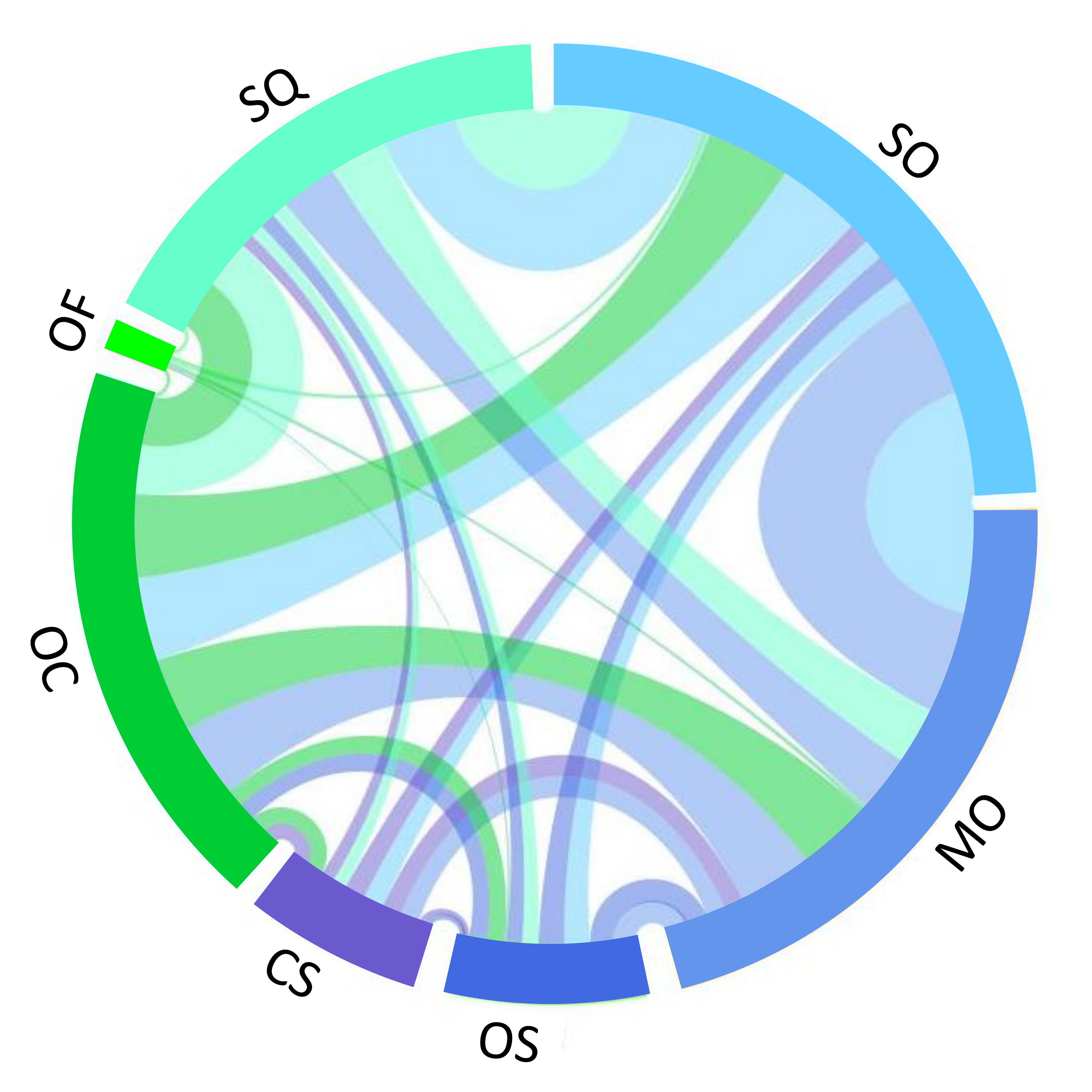}
\end{minipage}
}
\subfigure[Object Number $\&$ Size]{
\begin{minipage}[c]{0.52\linewidth}
\centering
\includegraphics[width=\textwidth]{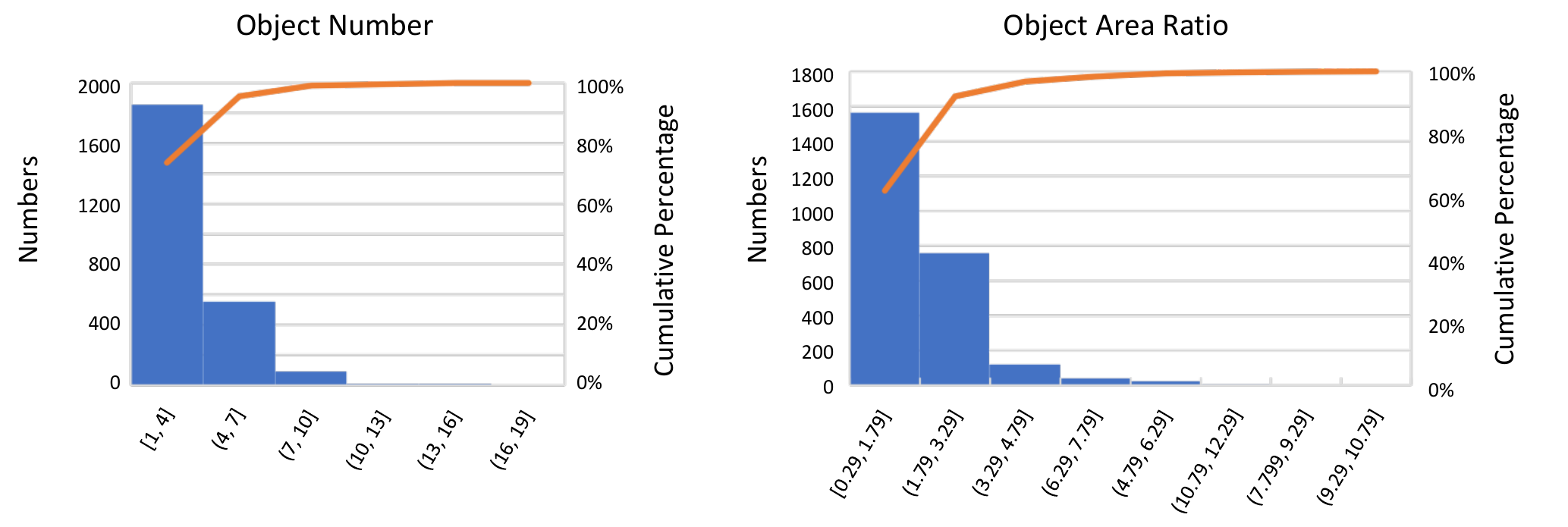}
\end{minipage}
}
\vspace{-10pt}
\caption{(a) Co-attribute distribution table (left) and multiple dependencies of these attributes (right). (b) Distribution of Trichomonas population in each image (left) and distribution of the object area ratio (right).
}
\label{Heatmap}
\end{figure*}

\vspace{-35pt}
\subsection{The Details of FBAs}
\vspace{-15pt}

\begin{figure}
    \centering
    \includegraphics[width=0.9\textwidth]{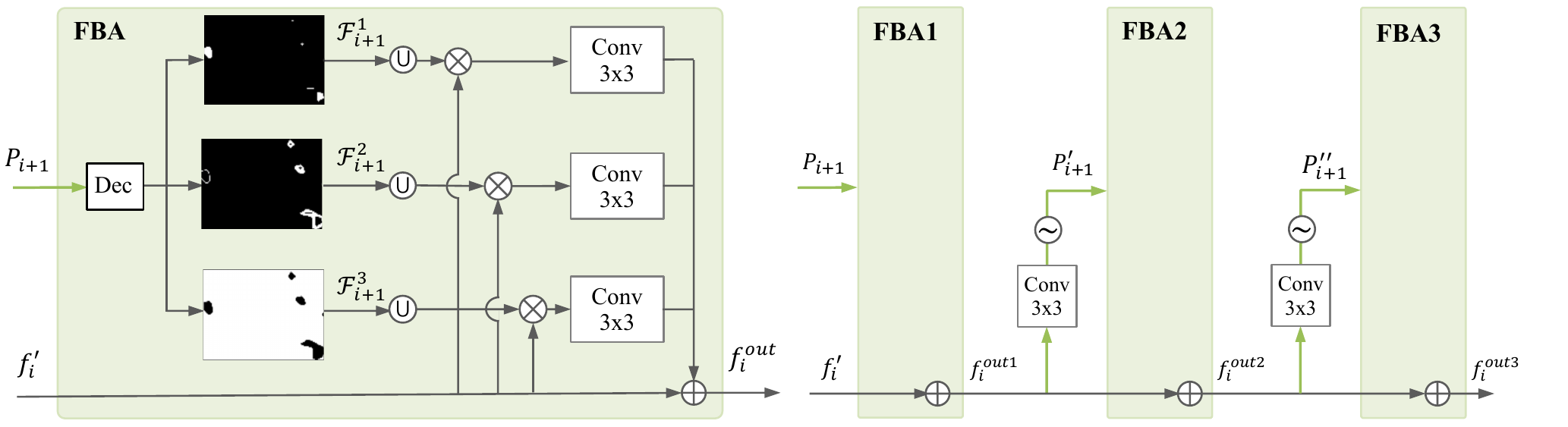}
    \vspace{-7pt}
    \caption{Cascaded form of multiple FBA modules. Left: Single FBA. Right: Cascaded FBAs. FBA can be cascaded multiple times to gradually refine predictions. ($\sim$): Sigmoid.
    }
    \label{fbas}
\end{figure}
\vspace{-5pt}

\vspace{-15pt}
\subsection{More Visual Comparisons}

\begin{figure}
    \centering
    \includegraphics[width=\textwidth]{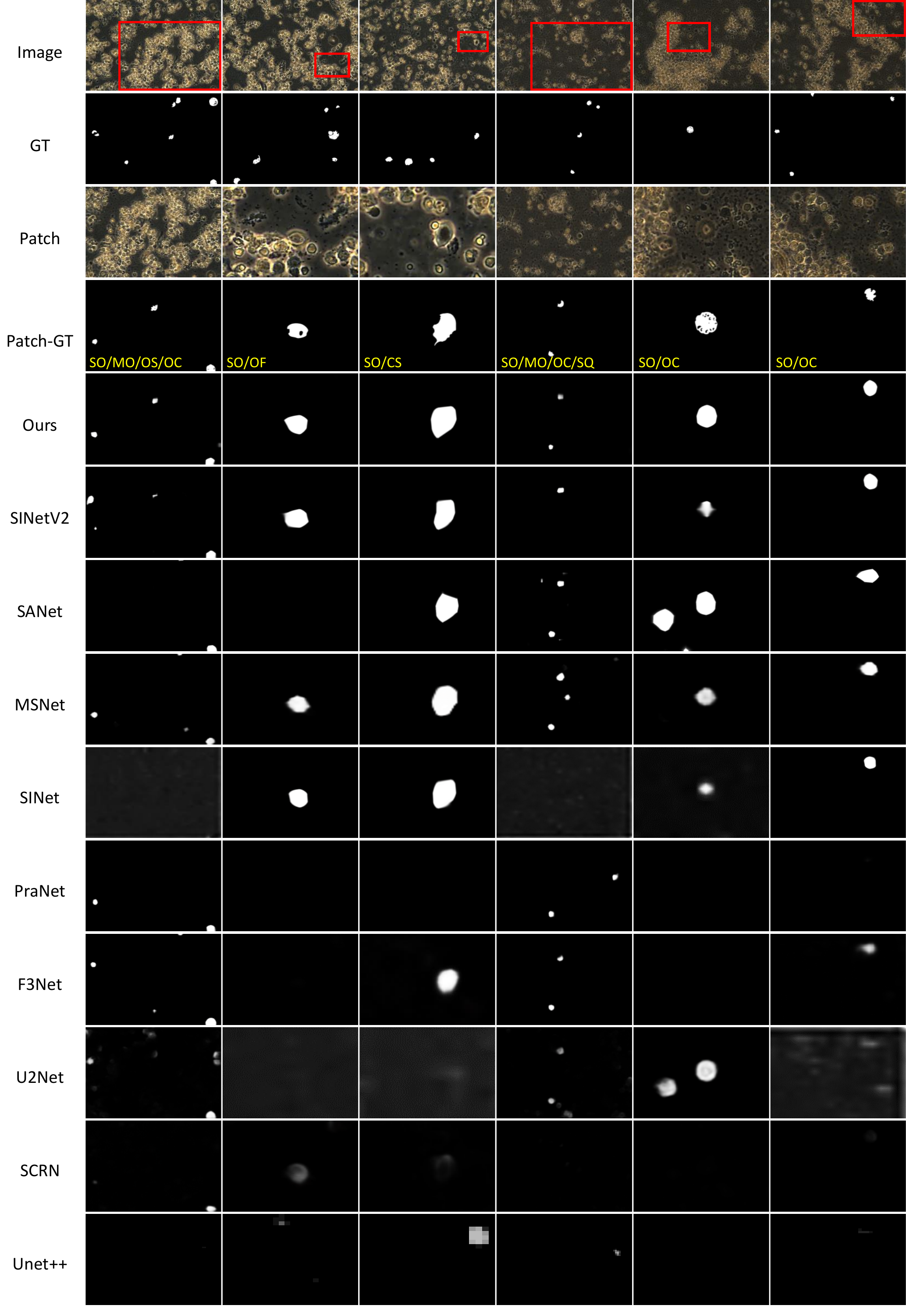}
    \caption{Visual comparison of different methods. We select image patches (red boxes) from input images to show the results more clearly (data attributes are also marked).}
    \label{RN}
\end{figure}

\end{document}